\documentclass[11pt]{article}
\usepackage[margin=1in]{geometry}
\usepackage{times}
\usepackage{graphicx}
\usepackage{amsmath}
\usepackage{amssymb}
\usepackage{comment}
\usepackage{hyperref}

\usepackage[numbers]{natbib}

\usepackage{subcaption}
\usepackage{booktabs}

\title{Scalable Data Attribution via Forward-Only Test-Time Inference}
\author{
    Sibo Ma \hspace{0.4cm} Julian Nyarko \\
    Stanford University \\
    \texttt{siboma@stanford.edu} \hspace{0.4cm} \texttt{jnyarko@stanford.edu}
}

\date{\today}

\begin{document}

\maketitle

\begin{abstract}
Data attribution seeks to trace model behavior back to the training examples that shaped it, enabling debugging, auditing, and data valuation at scale. 
Classical influence-function methods offer a principled foundation but remain impractical for modern networks because they require expensive backpropagation or Hessian inversion at inference. 
We propose a data attribution method that preserves the same first-order counterfactual target while eliminating per-query backward passes. 
Our approach simulates each training example’s parameter influence through short-horizon gradient propagation during training and later reads out attributions for any query using only forward evaluations. 
This design shifts computation from inference to simulation, reflecting real deployment regimes where a model may serve billions of user queries but originate from a fixed, finite set of data sources (for example, a large language model trained on diverse corpora while compensating a specific publisher such as the New York Times). 
Empirically, on standard MLP benchmarks, our estimator matches or surpasses state-of-the-art baselines such as TRAK on standard attribution metrics (LOO and LDS) while offering orders-of-magnitude lower inference cost. 
By combining influence-function fidelity with first-order scalability, our method provides a theoretical framework for practical, real-time data attribution in large pretrained models.
\end{abstract}

\section{Introduction}

Modern machine learning models inherit their behavior not only from their architectures but from the data on which they are trained. Understanding which training examples most strongly influence a given prediction, a task known as \textit{data attribution}, has become central to model transparency, auditing, and responsible deployment. 
Accurate attribution can identify mislabeled or poisoned data~\cite{koh2017understanding,northcutt2021confident}, trace the provenance of model outputs~\cite{ilyas2022datamodels,park2023trak}, and enable principled data valuation for collaborative or open-source training regimes~\cite{ghorbani2019data,jia2020efficient}. 
A central objective in these settings is to estimate how a counterfactual change in a single training example would alter a model’s prediction, a quantity formalized by influence functions as a first-order approximation to full retraining effects~\cite{koh2017understanding}. 
We build directly on this first-order counterfactual view, which we revisit in detail in Section~\ref{sec:method}.

Directly computing this quantity at inference time is infeasible for modern networks. 
Most influence estimators require either evaluating test-time gradients for every query $q$\cite{akyurek2022towards,park2023trak} or solving large inverse–Hessian systems\cite{koh2017understanding}. 
This imbalance can be problematic when inference dominates the overall usage of a model: once deployed, a pretrained foundation model may process millions or billions of queries, each potentially requiring attribution for auditing, revenue sharing, or compliance. 
In contrast, training, although expensive, happens only once or infrequently\cite{crankshaw2017clipperlowlatencyonlineprediction}. 
Thus, an ideal attribution system should shift computational effort from inference to training: we can afford heavy precomputation during model preparation but need extremely cheap, even forward-only, attribution during inference.

Such asymmetry arises naturally in many real-world scenarios with important policy implications.  
For instance, organizations seeking to compensate data contributors may precompute how each data source affects model parameters, then cheaply evaluate per-query royalties as the model is used\cite{wang2024economicsolutioncopyrightchallenges}.  
In large-scale generative models, even a single backpropagation per user query would be prohibitive; forward-only evaluation is a much more viable option.

We propose a forward-only data attribution method that preserves the classical influence function target while eliminating per-query backpropagation at test time.  
Our design explicitly shifts all heavy computation to training-time simulation: we perturb each training example’s contribution through short-horizon gradient propagation and store its resulting parameter response.  
At inference, attribution for a query $q$ reduces to two forward evaluations, under up- and down-weighted parameter states, yielding the same first-order effect as traditional influence functions.  
This design makes inference cost comparable to a standard prediction pass while maintaining attribution fidelity.  

Preliminary experiments show that our approach matches or exceeds the performance of recent baselines such as TRAK~\cite{park2023trak} on standard attribution metrics (LOO and LDS), while being significantly cheaper at test time.  
By combining the interpretability of influence functions with the scalability of forward-only evaluation, our method provides a practical and theoretically grounded mechanism designed to scale for tracing and valuing data in large-scale pretrained and fine-tuned models.

\section{Related work}

Training data attribution methods seek to quantify how individual training examples influence test-time predictions. 
The classical research on influence-functions defines this effect as the first-order counterfactual response to up- or down-weighting an example at the optimum, leading to the canonical form $-\,g_q^\top H^{-1} g_b$ \cite{koh2017understanding,agarwal2017second,martens2010deep,basu2021influenceg}. 
Variants accelerate computation by stochastic Hessian–vector inversion or damping \cite{schioppa2022scaling,agarwal2017second}, while trajectory-based methods such as TracIn estimate influence by accumulating gradient similarities along the training path \cite{pruthi2020estimating,charpiat2019input}. 
Representer-theorem approaches reinterpret predictions as linear combinations of training examples in regularized or last-layer-linear models \cite{yeh2018representer}, and recent data modeling frameworks learn such relationships directly by fitting predictive linear models over subsets \cite{ilyas2022datamodels} or through inner products between hidden activations and token-level prediction errors \cite{deng2025efficientforwardonlydatavaluation}. 
TRAK~\cite{park2023trak} bridges these perspectives through a linearized kernel approximation that computes leave-one-out effects via projected gradient features. 
Our method keeps the same IF target as these lines of work but achieves it through short-horizon gradient propagation, avoiding Hessian solves, stored trajectories, or explicit feature projection.

A complementary body of research studies how to approximate inverse curvature operators efficiently. 
Truncated Neumann expansions, stochastic recursions (LiSSA), and conjugate-gradient solvers provide unbiased or low-variance estimators of $H^{-1}g$ using Hessian–vector products \cite{martens2010deep,pearlmutter1994fast,agarwal2017second}. 
Alternative formulations replace $H$ with positive semi-definite surrogates such as the Gauss–Newton or Fisher matrix, further approximated through Kronecker or block factorizations (e.g., K-FAC, Shampoo) to scale to modern architectures \cite{martens2015optimizing,gupta2018shampoo,ba2017distributed}. 
More recent curvature-preconditioned optimizers exploit similar structures for efficient second-order adaptation \cite{anil2020scalable}. 
Our method is most closely related to Hessian–vector approaches, as it also targets the inverse–curvature product $H^{-1}g$, but it differs in where and how this effect is realized. 
Rather than explicitly solving for $H^{-1}g$ through iterative solvers or stochastic recursions at inference, we realize the same effect implicitly by locally simulating a few damped gradient updates around the trained model. 
These accumulated gradient updates pass the selected training examples’ gradient through the model’s local curvature and leave that response encoded in the resulting parameter displacement.
The resulting parameter displacement serves as a stored ``influence imprint'' that captures how the example would locally steer the model. 
At inference, this imprint is queried through forward-only evaluations, reproducing the influence-function response while avoiding per-query inverse Hessian–vector products or linear solves.

Another related line of research linearizes nonlinear networks around their converged weights to yield locally faithful models. 
The neural tangent kernel (NTK) and its empirical forms have been used to analyze optimization and generalization in overparameterized networks \cite{jacot2018neural,arora2019fine,bachmann2022generalization}, and several interpretability methods leverage gradient-feature embeddings for attribution or subpopulation discovery \cite{achille2021lqf,mohamadi2023fast,seleznova2022neural}. 
TRAK itself builds on this insight by replacing the network with its linear kernel approximation. 
Our approach shares the same linearization principle but applies it implicitly through finite-step propagation, producing a comparable curvature-aware effect while remaining fully model-agnostic and first-order in computation.

\section{Method}
\label{sec:method}
We present our procedure in three steps. \ref{sec:method_target} states the target (the standard IF quantity). \ref{sec:method_step2} demonstrates the perturbation of the training loss and how a single example moves the parameters. \ref{sec:method_step3} converts that parameter response into a forward-only test-time inference mechanism, which results in our proposed finite-difference readout $s(b, q)$ (see Equation \eqref{eq:fw-readout}, with cheaper variant in \eqref{eq:fw-readout-cheap}).

\subsection{Target Attribution Score}
\label{sec:method_target}
Our goal is to quantify how a single training example $b$ affects the model’s behavior (measured by F) on a query $q$. 
Following the classical influence function framework~\cite{koh2017understanding}, this effect can be expressed as a first-order counterfactual derivative of the model output with respect to the training weight of $b$.
Formally, we target the standard influence function (IF) quantity under the mean-loss convention:
\begin{equation}
\label{eq:if-target}
\tau_{\mathrm{IF}}(b\mid q) \;:=\; -\,\frac{1}{N}\, g_q^\top\, H^{-1}\, g_b,
\end{equation}
where $H=\nabla_\theta^2 L(\theta^*)$ is the Hessian of the total training loss at the converged parameters $\theta^*$ and $N$ is the number of training examples. 
Throughout, we define $g_b := \nabla_\theta \ell(\theta^*; b)$ and $g_q := \nabla_\theta F(q;\theta^*)$ as the loss and query gradients, respectively.

Intuitively, \eqref{eq:if-target} measures the sensitivity of the model output on $q$ to an infinitesimal reweighting of $b$. 
Computing this exactly requires solving a linear system involving $H^{-1}$, which is infeasible for modern networks. 
The remainder of our method develops a forward-simulation procedure that reproduces this same first-order response without explicitly inverting the Hessian.

\subsection{Simulating a training example’s influence}
\label{sec:method_step2}
To avoid directly computing $H^{-1}$, we simulate how reweighting a single example would perturb the training dynamics. 
The idea is to inject a small targeted change in the loss function, follow its evolution under gradient descent, and then read out the resulting parameter displacement. 
This procedure effectively constructs an implicit approximation of $H^{-1} g_b$ using only forward steps.

\subsubsection*{Setup}
We start from the empirical objective
\begin{equation*}
    L(\theta)\;=\;\frac{1}{N}\sum_{i=1}^N \ell(\theta;b_i),
\end{equation*}
and consider its up- and down-weighted versions for a specific example $b$:
\begin{equation}
L_{\pm\varepsilon}(\theta) \;=\; L(\theta) \;\pm\; \frac{\varepsilon}{N}\,\ell(\theta;b).
\label{eq:updown}
\end{equation}
Running $T$ steps of gradient descent on these perturbed objectives from the reference point $\theta^*$ with step size $\eta>0$ gives
\begin{equation}
\theta_{t+1}^{(\pm)} \;=\; \theta_t^{(\pm)} - \eta\,\nabla L_{\pm\varepsilon}\!\big(\theta_t^{(\pm)}\big).
\label{eq:gdplusminus}
\end{equation}
Let $\Delta\theta_t^{(\pm)} := \theta_t^{(\pm)}-\theta^*$, and define the antisymmetric and symmetric displacements
\[
\Delta\delta_t:=\Delta\theta_t^+-\Delta\theta_t^-,\qquad
\Delta\Sigma_t:=\Delta\theta_t^++\Delta\theta_t^-.
\]
By construction, $\Delta\delta_t$ captures the first-order effect of reweighting $b$, while $\Delta\Sigma_t$ collects the higher-order residuals that we will later bound.

\subsubsection*{Linearized updates}
Because we study a trained model, the gradient of the mean loss is near zero, $\nabla L(\theta^*)\approx 0$. 
For sufficiently small perturbations $\varepsilon$, a first-order Taylor expansion around $\theta^*$ yields (see Appendix~\ref{appx:step2-taylor})
\begin{equation}
\Delta\delta_{t+1} \;=\; (I-\eta H)\,\Delta\delta_t \;-2 \eta\,\frac{\varepsilon}{N}\,g_b\;+\; O\!\big((\varepsilon/N)^3\big),
\label{eq:firstorder-s}
\end{equation}
and
\begin{equation}
\Delta\theta^\pm_{t+1} \;=\; (I-\eta H)\,\Delta\theta^\pm_t \;\mp\eta\,\frac{\varepsilon}{N}\,g_b\;+\; O\!\big((\varepsilon/N)^2\big).
\label{eq:firstorder-s-half}
\end{equation}
These recursions describe how a small perturbation to one example propagates through iterative optimization.

\subsubsection*{Unrolling the response}
To make the dependence on $H$ explicit, define $A:=I-\eta H$ and $u:=-2\,\eta\,\frac{\varepsilon}{N}\,g_b+O\!\big((\varepsilon/N)^3\big)$. 
Unrolling the recursion \eqref{eq:firstorder-s} over $T$ steps (see Appendix~\ref{appx:step2-unroll}) gives
\[
\Delta\delta_T \;=\; u\sum_{t=0}^{T-1} A^t \;=\; -2\frac{\varepsilon}{N}\,\Big(\eta\sum_{t=0}^{T-1} (I-\eta H)^t\Big)\,g_b+ O\!\big((\varepsilon/N)^3\big).
\]
We define the cumulative response operator
\[
S_T(H)\;:=\;\eta\sum_{t=0}^{T-1}(I-\eta H)^t,
\]
so that
\begin{equation}
    \Delta\delta_T \;=\; -2\frac{\varepsilon}{N}\,S_T(H)\,g_b+ O\!\big((\varepsilon/N)^3\big).
\label{eq:deltaT-upw}
\end{equation}
This expression summarizes how the perturbation accumulates over training: $\Delta\delta_T$ directly links the example gradient $g_b$ to the resulting parameter displacement through the linear operator $S_T(H)$.

\subsubsection*{Damping and the $T\!\to\!\infty$ limit}
In deep networks, the Hessian $H$ can have negative curvature directions, which make the iteration unstable. 
We therefore add Tikhonov damping via $H_\lambda:=H+\lambda I$ and choose $\lambda,\eta$ so that $\|I-\eta H_\lambda\|<1$. 
Under this condition,
\[
S_T(H_\lambda) = \eta\sum_{t=0}^{T-1} (I-\eta H_\lambda)^t = H_\lambda^{-1}\!\left(I-(I-\eta H_\lambda)^T\right)
\xrightarrow[T\to\infty]{} H_\lambda^{-1}.
\]
Hence, as $T\!\to\!\infty$,
\[
\Delta\delta_T\;=\;-2\frac{\varepsilon}{N}\,H_\lambda^{-1}\,g_b+ O\!\big((\varepsilon/N)^3\big),
\qquad
\Delta\theta_T^\pm\;=\;\mp\frac{\varepsilon}{N}\,H_\lambda^{-1}\,g_b+ O\!\big((\varepsilon/N)^2\big).
\]
This result completes Step~2: through a forward simulation, we have recovered a stable, differentiable estimate of the parameter shift corresponding to up- or down-weighting $b$.

\subsection{From parameter response to an influence score on $q$}
\label{sec:method_step3}
Having obtained the parameter displacement caused by reweighting $b$, we now translate it into an observable effect on a test query $q$. 
The key idea is simple: if we evaluate the model output $F(q;\theta)$ under the two perturbed parameters $\theta^*+\Delta\theta^\pm_T$, their difference isolates the same first-order term that appears in the influence function target \eqref{eq:if-target}.

We define the symmetric finite-difference readout:
\begin{equation}
s(b, q)
\;:=\;
\,\frac{F\!\big(q;\theta^*+\Delta\theta^+_T(b;\varepsilon)\big)\;-\;F\!\big(q;\theta^*+\Delta\theta^-_T(b;\varepsilon)\big)}{2\,(\varepsilon/N)}.
\label{eq:fw-readout}
\end{equation}
This forward-only computation is conceptually analogous to probing the model with a virtual “up-weight” and “down-weight” of $b$ and reading out the resulting change in prediction on $q$.

Let $\Delta^\pm := \Delta\theta^\pm_T(b;\varepsilon)$ and $H_q:=\nabla_\theta^2 F(q;\theta^*)$. 
A second-order Taylor expansion of $F(q;\cdot)$ at $\theta^*$ (Appendix~\ref{appx:step3-taylor}) yields
\[
F\big(q;\theta^*+\Delta^+\big)-F\big(q;\theta^*+\Delta^-\big)
= g_q^\top\Delta\delta_T \;+\; \frac{1}{2}\,\Delta\delta_T^\top H_q\,\Delta\Sigma_T \;+\; O\!\big(\|\Delta^+\|^3+\|\Delta^-\|^3\big).
\]
Because $\Delta\delta_T=O(\varepsilon/N)$ and $\Delta\Sigma_T=O((\varepsilon/N)^2)$, higher-order terms vanish, leaving
\[
F\big(q;\theta^*+\Delta^+\big)-F\big(q;\theta^*+\Delta^-\big)
= g_q^\top\Delta\delta_T \;+\; O\!\big((\varepsilon/N)^3\big).
\]
Substituting this into \eqref{eq:fw-readout} and using \eqref{eq:deltaT-upw} gives
\begin{equation}
s(b, q)
= -\,g_q^\top S_T(H_\lambda)\,g_b\;+\;O\!\big((\varepsilon/N)^2\big)
\;\xrightarrow[T\to\infty]{}\;
-\,g_q^\top H_\lambda^{-1} g_b\;+\;O\!\big((\varepsilon/N)^2\big).
\end{equation}
Thus, the simple difference of two forward evaluations reproduces the canonical influence function while remaining free of any backward Hessian inversion.

\subsubsection*{A cheaper variant (half the compute; higher bias)}
For additional efficiency, we can mirror a single trajectory rather than compute two. 
Define
\begin{equation}
s(b, q)
\;:=\;
\,\frac{F\!\big(q;\theta^*+\Delta\theta_T^+(b;\varepsilon)\big)\;-\;F\!\big(q;\theta^*-\Delta\theta_T^+(b;\varepsilon)\big)}{2\,(\varepsilon/N)}.
\label{eq:fw-readout-cheap}
\end{equation}
This reduces compute by half while introducing only higher-order residual bias:
\begin{equation}
s(b, q)
= -\,g_q^\top S_T(H_\lambda)\,g_b\;+\;O\!\big(\varepsilon/N\big)
\;\xrightarrow[T\to\infty]{}\;
-\,g_q^\top H_\lambda^{-1} g_b\;+\;O\!\big(\varepsilon/N\big).
\end{equation}
Empirically, this “single-trajectory” estimator trades a small increase in bias for a significant computational saving, making it useful in large-scale settings.

\section{Evaluation}

We evaluate our method using the \textsc{Dattri} framework~\cite{NEURIPS2024_f7326833}, which provides a unified protocol for benchmarking data attribution models. Our evaluation proceeds in three parts.  
First, we analyze the effects of the influencial hyperparameters of our method: unroll depth $T$, step size $\eta$, and damping $\lambda$.  
Second, we compare our method’s performance against existing attribution methods.  
Finally, we analyze the time and memory efficiency relative to prior influence-function approximations.
Attribution quality is measured using two established metrics implemented in the \textsc{Dattri} benchmark~\cite{NEURIPS2024_f7326833}:

\begin{description}
    \item[Leave-one-out (LOO)\cite{koh2017understanding}] 
    Quantifies how well an attribution method reproduces the true counterfactual effect of individual training examples.  
    It measures the correlation between predicted influence scores and the actual change in model output when each training point is removed and the model is retrained.  
    Higher LOO correlation indicates closer agreement with exact retraining outcomes.
    
    \item[Linear Datamodeling Score (LDS)\cite{ilyas2022datamodels,park2023trak}] 
    Measures how well attributions linearly predict model outputs across many random training subsets. 
    It computes the correlation between predicted subset-level outputs (from summed per-example scores) and those obtained from actual retrained models. 
    A higher LDS means the method consistently models additive data effects across subsets.
\end{description}
Both metrics are standard in recent data-attribution studies and together form a reliable protocol for comparing methods across architectures and datasets.

\subsection{Hyperparameter analysis}
We examine how unroll depth $T$, step size $\eta$, and damping $\lambda$ influence attribution accuracy.  
These parameters determine how the forward simulation approximates the damped influence-function target.  
The estimation error, defined as the difference between our simulated estimate and the exact target, can be expressed as

\[
-\,g_q^\top S_T(H_\lambda) g_b
+ g_q^\top H_\lambda^{-1} g_b
= g_q^\top (I - \eta H_\lambda)^{T} H_\lambda^{-1} g_b.
\]
Therefore, the error depends on how the hyperparameters shape the term $(I - \eta H_\lambda)^T$.

\paragraph{Unroll depth $T$.}
The unroll depth determines how many simulated gradient steps are performed and thus how completely curvature effects are captured.  
When $T$ is small, the approximation is truncated early and misses part of the curvature contribution.  
As predicted by the geometric decay of $A^T$, performance improves rapidly at first and then levels off as $\|A^T\|$ becomes small.  
In Figure~\ref{fig:LDS_LOO_T}, both LOO and LDS scores increase sharply when $T<200$ and plateau thereafter, confirming this pattern.  
For very large $T$, performance slightly decreases for the largest $\eta$, suggesting that overly long unrolling can move the simulation outside the local region where the linearized curvature approximation defined by $H_\lambda$ holds.

\paragraph{Inner step size $\eta$.}
The step size $\eta$ controls how far each simulated update moves and how quickly $A^T$ shrinks.  
A small $\eta$ leads to slow progress and delayed convergence, while a large $\eta$ risks instability because some eigenvalues of $A$ may leave the unit circle, violating the condition $\|I - \eta H_\lambda\| < 1$.  
Figure~\ref{fig:LDS_LOO_T} shows that a moderate step size ($\eta = 3\times10^{-2}$) achieves the most stable and monotonic improvement, whereas smaller $\eta$ converges slowly and larger $\eta$ achieves faster early gains but eventually declines due to instability.

\paragraph{Damping $\lambda$.}
Damping modifies the curvature matrix as $H_\lambda = H + \lambda I$, which increases all eigenvalues and helps satisfy $\|I - \eta H_\lambda\| < 1$.  
This shift makes it easier to satisfy the stability condition $\|I - \eta H_\lambda\| < 1$, ensuring geometric convergence of the error term.  
When $\lambda$ is large, the system becomes more stable and less sensitive to poorly conditioned directions in $H$, but introduces bias because it targets the damped inverse $H_\lambda^{-1}$ rather than the true inverse $H^{-1}$. 
A smaller $\lambda$ reduces this bias but can make convergence slower or unstable when $H$ contains very small or negative eigenvalues that violate the stability condition.

\subsection{Performance comparison with existing works}

\begin{figure}[t]
    \centering
    \begin{subfigure}[t]{0.48\linewidth}
        \centering
        \includegraphics[width=\linewidth]{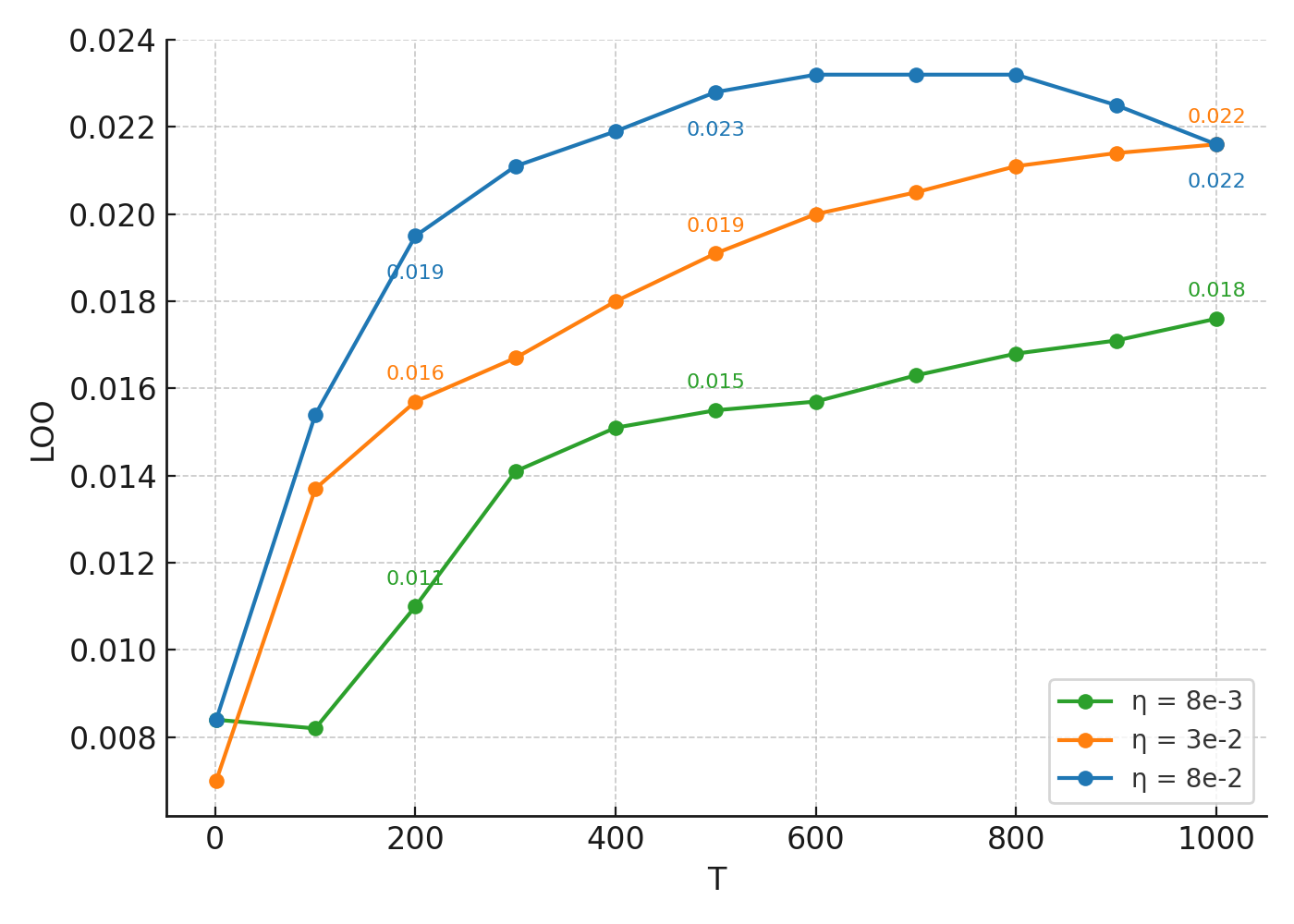}
        \caption{Performance on LOO under different $\eta$ and $T$.}
        \label{fig:LOO_vs_T}
    \end{subfigure}%
    \hfill
    \begin{subfigure}[t]{0.48\linewidth}
        \centering
        \includegraphics[width=\linewidth]{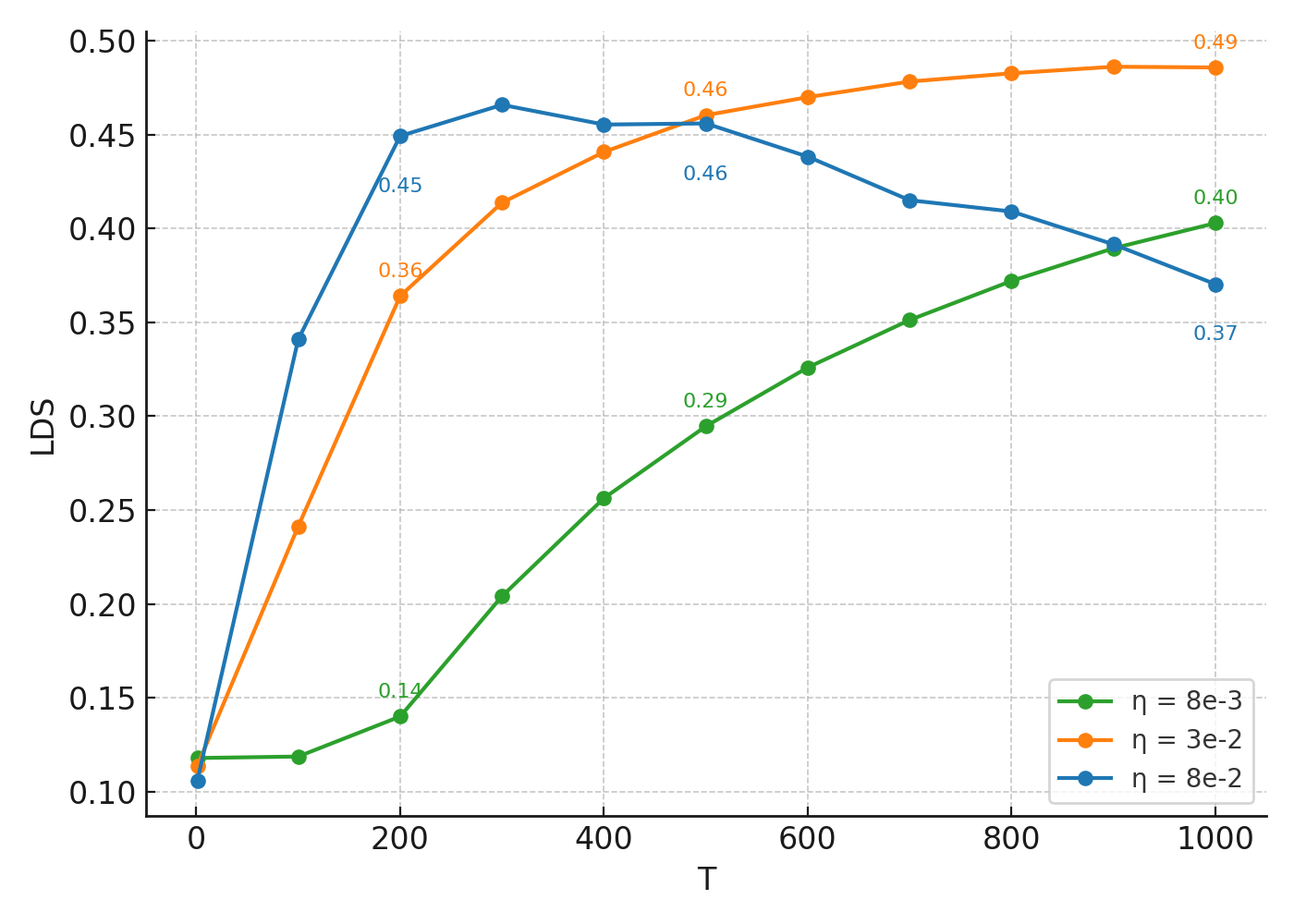}
        \caption{Performance on LDS under different $\eta$ and $T$.}
        \label{fig:LDS_vs_T}
    \end{subfigure}
    \caption{Comparison of LOO and LDS performance under varying $\eta$ and unroll steps $T$. 
    On a small MLP trained on MNIST, a moderate $\eta$ ($3\times10^{-2}$) yields stable and monotonic improvement. 
    Larger $\eta$ accelerates early gains but leads to instability, while smaller $\eta$ converges slowly yet remains stable.}
    \label{fig:LDS_LOO_T}
\end{figure}

\begin{figure}[th]
    \centering
    \begin{subfigure}[t]{0.48\linewidth}
        \centering
        \includegraphics[width=\linewidth]{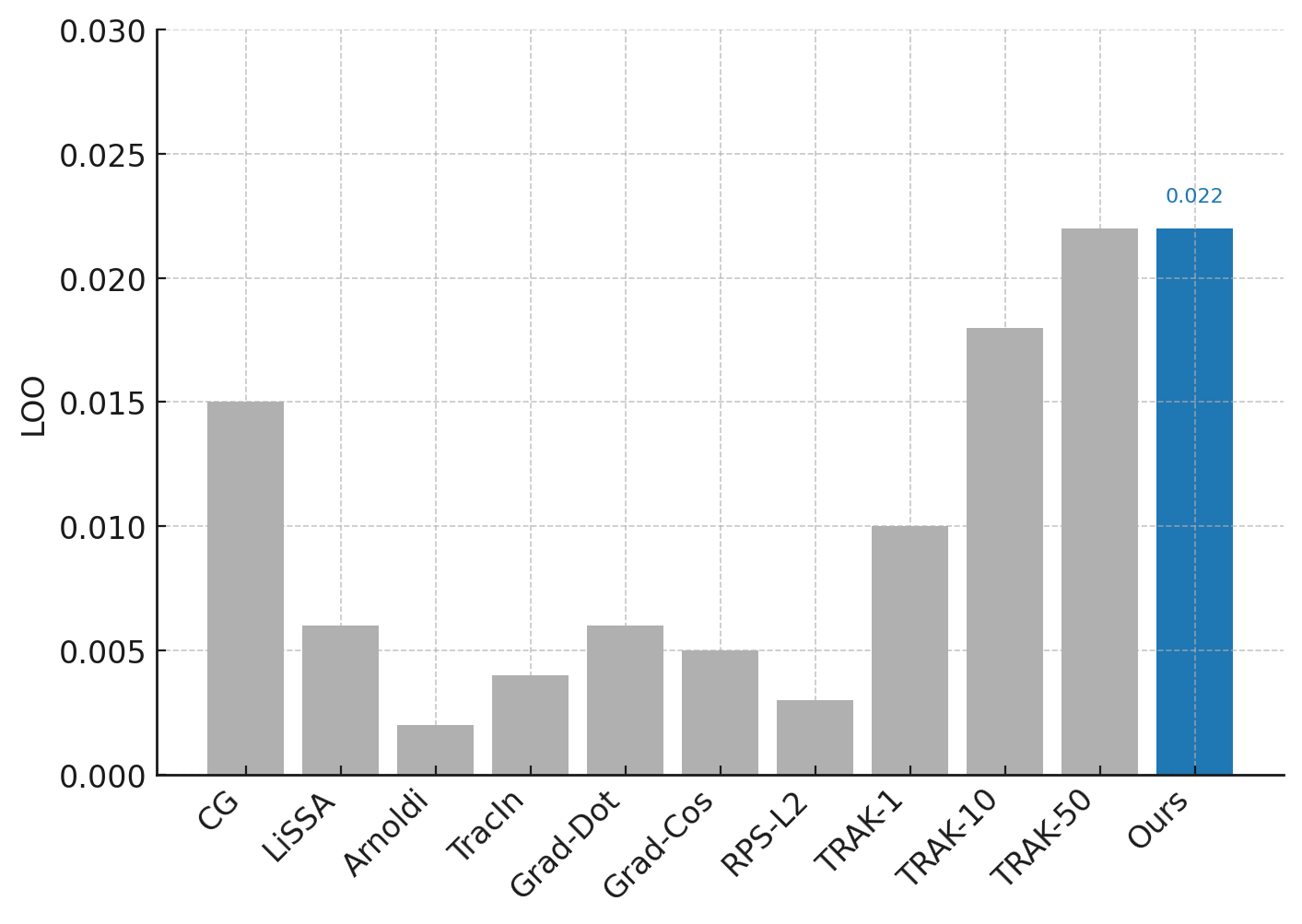}
        \caption{LOO correlation of different attribution methods on MNIST–MLP.}
        \label{fig:LOO_vs_T_other}
    \end{subfigure}%
    \hfill
    \begin{subfigure}[t]{0.48\linewidth}
        \centering
        \includegraphics[width=\linewidth]{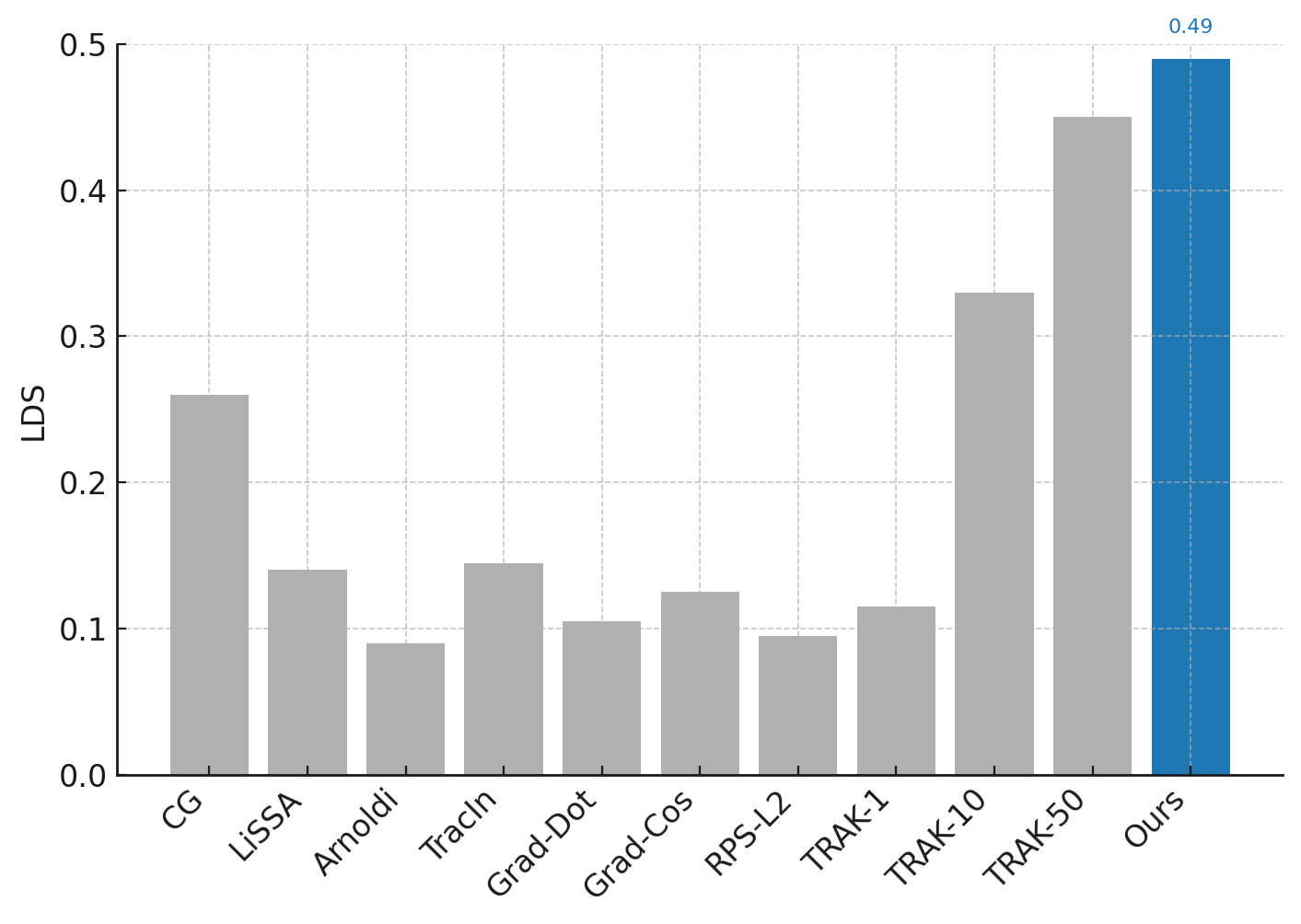}
        \caption{LDS correlation of different attribution methods on MNIST–MLP.}
        \label{fig:LDS_vs_T_other}
    \end{subfigure}
    \caption{Performance comparison on the MNIST–MLP benchmark from \textsc{Dattri}~\cite{NEURIPS2024_f7326833}. 
Each bar reports the correlation under LOO and LDS. 
Our method achieves $0.022$ LOO and $0.49$ LDS, matching or exceeding the strongest baselines while requiring only forward evaluations at inference time.}

    \label{fig:LDS_LOO_comparison}
\end{figure}

We benchmark our estimator on MNIST using an MLP with 0.11M parameters.  
This compact setting allows precise measurement of convergence trends and facilitates direct comparison across $\eta$ and $T$.  
Figure~\ref{fig:LDS_LOO_T} shows how both LOO and LDS metrics evolve as we vary these parameters.  
Our method attains LOO scores around $0.02$ and LDS scores near $0.49$, comparable to the best-performing baselines in the \textsc{Dattri} framework.  
As expected, moderate step sizes balance accuracy and stability, while extreme $\eta$ values trade one for the other.

Figure~\ref{fig:LDS_LOO_comparison} reproduces the benchmark results from \textsc{Dattri}~\cite{NEURIPS2024_f7326833}, summarizing the leave-one-out (LOO) and linear datamodeling score (LDS) performance of existing attribution methods across larger architectures. 
Our method achieves the highest LDS and LOO scores among all baselines, indicating stronger alignment with the true influence-function targets. 

\subsection{Analytical comparison of computational complexity}

Finally, we provide an analytical comparison of computational cost and memory footprint relative to representative attribution methods.  
We separate total complexity into (i) simulation cost during the forward unrolling phase and (ii) inference cost during test-time readout.

\paragraph{Simulation cost.}
Our method performs $T$ gradient updates for each selected training example.  
If $|B|$ denotes the number of examples whose influence is simulated, the total cost scales as $O(|B|T)$ backpropagations.  
This is linear in the unroll depth and independent of the total dataset size when $|B|\!\ll\!N$.

\paragraph{Inference cost.}
At test time, each $(q,b)$ pair requires two forward passes—one for the up-weighted and one for the down-weighted trajectory—yielding a total cost of $O(2|Q||B|)$ forward evaluations.  
This is the same order as other forward-only influence estimators and orders of magnitude cheaper than methods requiring repeated model retraining, explicit Hessian inversion, or per-sample gradient calculation.

\bibliographystyle{plainnat}
\bibliography{reference}

\clearpage
\appendix
\section{Mathematical Proofs}
This appendix provides the detailed derivations referenced in Section \ref{sec:method_target}. 
Each subsection mirrors the corresponding step in the main text and uses identical notation and labels for ease of cross-reference.

\subsection{Step 2: Taylor expansion and induction}
\label{appx:step2-taylor}
This subsection formalizes the first-order expansion used in Step~2 of the main text. 
We show that the displacement of the model parameters under the up- and down-weighted objectives evolves linearly to first order, and we establish the scaling bounds on $\Delta\delta_t$ and $\Delta\Sigma_t$.

\vspace{0.3em}
\noindent\textbf{Setup.}  
Let $\Delta\theta_t^{(\pm)} := \theta_t^{(\pm)}-\theta^*$. 
Subtracting $\theta^*$ from both sides of the update rule~\eqref{eq:gdplusminus} and substituting the perturbed objectives~\eqref{eq:updown} yields
\[
\Delta\theta_{t+1}^{(\pm)}
=\Delta\theta_t^{(\pm)}-\eta\!\left(\nabla L(\theta_t^{(\pm)})\pm\frac{\varepsilon}{N}\,\nabla\ell(\theta_t^{(\pm)};b)\right).
\]

\vspace{0.3em}
\noindent\textbf{Local expansion.}
Applying a first-order Taylor expansion around $\theta^*$ for both gradient terms gives
\begin{align*}
\nabla L(\theta_t^{(\pm)}) &=\nabla L(\theta^*+\Delta\theta_t^{(\pm)})
= \nabla L(\theta^*) + H\,\Delta\theta_t^{(\pm)} + O\!\big(\|\Delta\theta_t^{(\pm)}\|^2\big),\\
\nabla\ell(\theta_t^{(\pm)};b) &= \nabla\ell(\theta^*+\Delta\theta_t^{(\pm)};b)
= g_b + H_b\,\Delta\theta_t^{(\pm)} + O\!\big(\|\Delta\theta_t^{(\pm)}\|^2\big),
\end{align*}
where $H$ and $H_b$ denote the Hessians of the mean loss and the per-example loss, respectively. 
Because the model is trained to convergence, $\nabla L(\theta^*)\approx 0$.

Plugging these expansions into the update above and discarding negligible higher-order terms yields
\begin{align*}
\Delta\theta_{t+1}^{(\pm)}
&=(I-\eta H)\Delta\theta_t^{(\pm)}\mp\eta\,\frac{\varepsilon}{N}\,g_b
\;\mp\;\eta\,\frac{\varepsilon}{N}\,H_b\Delta\theta_t^{(\pm)}
\;+\; O((\varepsilon/N)^2) \;\mp\; O((\varepsilon/N)^3).
\end{align*}

\vspace{0.3em}
\noindent\textbf{Combining symmetric and antisymmetric components.}
To simplify the coupled recursions, define 
\[
\Delta\delta_t := \Delta\theta_t^+ - \Delta\theta_t^-, 
\qquad 
\Delta\Sigma_t := \Delta\theta_t^+ + \Delta\theta_t^-.
\]
Subtracting and adding the $\pm$ updates gives
\begin{align*}
\Delta\delta_{t+1}
&=(I-\eta H)\Delta\delta_t - 2\eta\,\frac{\varepsilon}{N}\,g_b
  - \eta\,\frac{\varepsilon}{N}\,H_b\Delta\Sigma_t + O((\varepsilon/N)^3),\\[0.4em]
\Delta\Sigma_{t+1}
&=(I-\eta H)\Delta\Sigma_t - \eta\,\frac{\varepsilon}{N}\,H_b\Delta\delta_t + O((\varepsilon/N)^2).
\end{align*}

\vspace{0.3em}
\noindent\textbf{Scaling argument.}
By induction on $t$, one can verify that if $\Delta\delta_t=O(\varepsilon/N)$ and $\Delta\Sigma_t=O((\varepsilon/N)^2)$, the same bounds hold at step $t{+}1$. 
Since $\Delta\delta_0=\Delta\Sigma_0=0$, these orders of magnitude persist throughout training. 
Hence, the first-order terms dominate and the linearized update~\eqref{eq:firstorder-s} in the main text is justified.
\subsection{Step 2: Unrolling the response}
\label{appx:step2-unroll}
We now solve the linear recurrence obtained above to express $\Delta\delta_T$ in closed form. 
Ignoring higher-order corrections, the recurrence \eqref{eq:firstorder-s} can be written compactly as
\[
\Delta\delta_{t+1}=A\,\Delta\delta_t+u,\qquad 
A:=I-\eta H, \quad u:=-2\,\eta\,\frac{\varepsilon}{N}\,g_b+O((\varepsilon/N)^3),
\]
with initialization $\Delta\delta_0=0$.

Iterating this recursion yields
\[
\Delta\delta_T \;=\; \sum_{k=0}^{T-1} A^k\,u
\;=\; -2\,\frac{\varepsilon}{N}\,\Big(\eta\sum_{k=0}^{T-1} (I-\eta H)^k\Big)\,g_b 
\;+\; O((\varepsilon/N)^3).
\]
When the spectral radius $\|A\|<1$ (ensured by the damping condition on $\eta$ and $\lambda$), the geometric series is uniformly bounded,
\[
\Big\|\sum_{k=0}^{T-1}A^k\Big\|\le \frac{1-\|A\|^T}{1-\|A\|}=O(1),
\]
so the accumulated remainder remains of order $O((\varepsilon/N)^3)$.

\subsection{Step 3: Taylor expansion for the forward readout}
\label{appx:step3-taylor}
Finally, we expand the test function $F(q;\theta)$ to justify the readout formula in~\eqref{eq:fw-readout}. 
Let
\[
\Delta^+ := \Delta\theta^+_T(b;\varepsilon),\qquad
\Delta^- := \Delta\theta^-_T(b;\varepsilon),\qquad
H_q:=\nabla_\theta^2 F(q;\theta^*).
\]
Applying a second-order Taylor expansion of $F(q;\cdot)$ at $\theta^*$ gives
\begin{align*}
F(q;\theta^*+\Delta^+)
&= F(q;\theta^*) + g_q^\top \Delta^+
   + \tfrac{1}{2}\,(\Delta^+)^\top H_q\,\Delta^+ + O(\|\Delta^+\|^3),\\
F(q;\theta^*+\Delta^-)
&= F(q;\theta^*) + g_q^\top \Delta^-
   + \tfrac{1}{2}\,(\Delta^-)^\top H_q\,\Delta^- + O(\|\Delta^-\|^3).
\end{align*}
Since $H_q$ is symmetric, the cross terms cancel:
\[
\Delta\delta_T^\top H_q\,\Delta\Sigma_T
=(\Delta^+-\Delta^-)^\top H_q(\Delta^++\Delta^-)
=(\Delta^+)^\top H_q\Delta^+-(\Delta^-)^\top H_q\Delta^-.
\]
Subtracting the two expansions thus yields
\[
F(q;\theta^*+\Delta^+)-F(q;\theta^*+\Delta^-)
= g_q^\top\Delta\delta_T 
  + \tfrac{1}{2}\,\Delta\delta_T^\top H_q\,\Delta\Sigma_T
  + O(\|\Delta^+\|^3+\|\Delta^-\|^3).
\]
Because $\Delta\delta_T=O(\varepsilon/N)$ and $\Delta\Sigma_T=O((\varepsilon/N)^2)$, the quadratic and cubic terms are negligible, confirming the approximation used in Step~3 of the main text.
\end{document}